\documentclass[journal]{IEEEtran}

\usepackage{graphicx,subcaption}
\usepackage{amssymb}
\usepackage{makecell}
\usepackage[dvipsnames]{xcolor}
\graphicspath{{./images/}}
\usepackage{multirow}
\usepackage{amsmath}
\usepackage{enumerate}
\usepackage{pifont}
\newcommand{\ie}{{\it i.e.}}
\newcommand{\eg}{{\it e.g.}}
\newcommand{\wrt}{{\it w.r.t.}}
\newcommand{\etc}{{\it etc.}}
\newcommand{\cmark}{\ding{51}}%
\newcommand{\xmark}{\ding{55}}%

\begin{document}

\title{Make Skeleton-based Action Recognition Model \\Smaller, Faster and Better}

\author{Fan~Yang$^{1,2}$,
        Sakriani~Sakti$^{1,2}$,
        Yang~Wu$^{3}$,
        ~Satoshi~Nakamura$^{1,2}$
        \newline
        $^1$ Nara Institute of Science and Technology, Japan~
        $^2$ RIKEN, AIP, Japan~
        $^3$ Kyoto University, Japan}


\maketitle

\begin{abstract}
Although skeleton-based action recognition has achieved great success in recent years, most of the existing methods may suffer from a large model size and slow execution speed. To alleviate this issue, we analyze skeleton sequence properties to propose a Double-feature Double-motion Network (DD-Net) for skeleton-based action recognition. By using a lightweight network structure (\ie,~ 0.15 million parameters), DD-Net can reach a super fast speed, as 3,500 FPS on one GPU, or, 2,000 FPS on one CPU. By employing robust features, DD-Net achieves the state-of-the-art performance on our experiment datasets: SHREC (\ie,~ hand actions) and JHMDB (\ie,~body actions). Our code is on https://github.com/fandulu/DD-Net.
\end{abstract}


\IEEEpeerreviewmaketitle

\section{Introduction}

Skeleton-based action recognition has been widely used in multimedia applications, such as human-computer interaction~\cite{ren2011robust}, human behavior understanding~\cite{wei2014skeleton} and medical assistive applications~\cite{chang2011kinect}. However, most of the existing methods may suffer from a large model size and slow execution speed~\cite{de2017shrec, devineau2018convolutional, hou2018spatial, zolfaghari2017chained, choutas2018potion}. 

In real applications, a desirable skeleton-based action recognition model should run efficiently by using a few parameters, and, also be adaptable to various application scenarios (\eg,~ hand/body, 2D/3D skeleton, and actions related/unrelated to global trajectories). To achieve this goal, we investigate skeleton sequence properties to propose a lightweight Double-feature Double-motion Network (DD-Net), which is equipped with a Joint Collection Distances (JCD) feature and a two-scale global motion feature.

\begin{figure}[ht!]
\centering
\begin{subfigure}[t]{.9\linewidth}
  {\includegraphics[width=.9\linewidth]{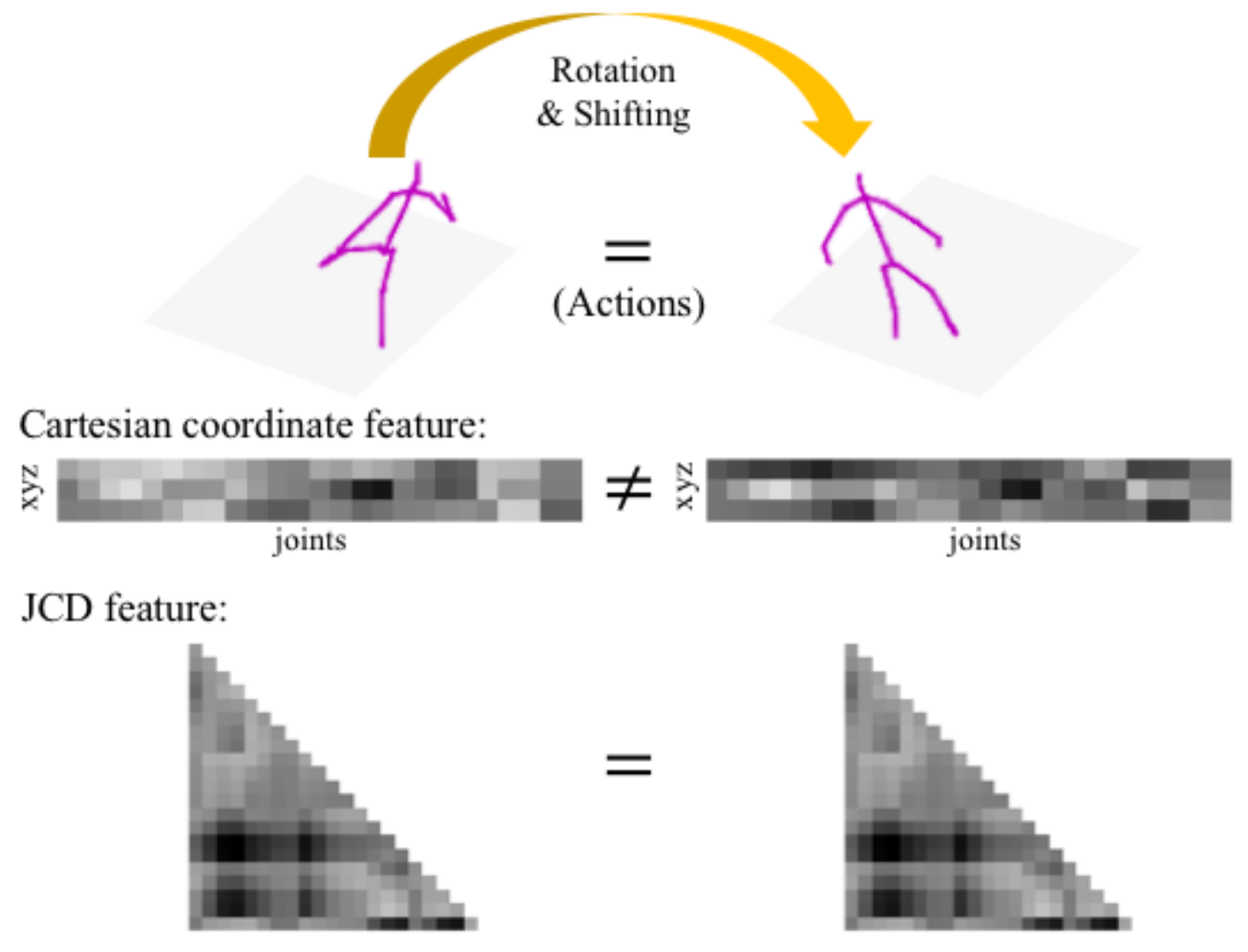}}
   \caption{Location-viewpoint variation} 
\end{subfigure}
\begin{subfigure}[t]{.9\linewidth}
  {\includegraphics[width=.8\linewidth]{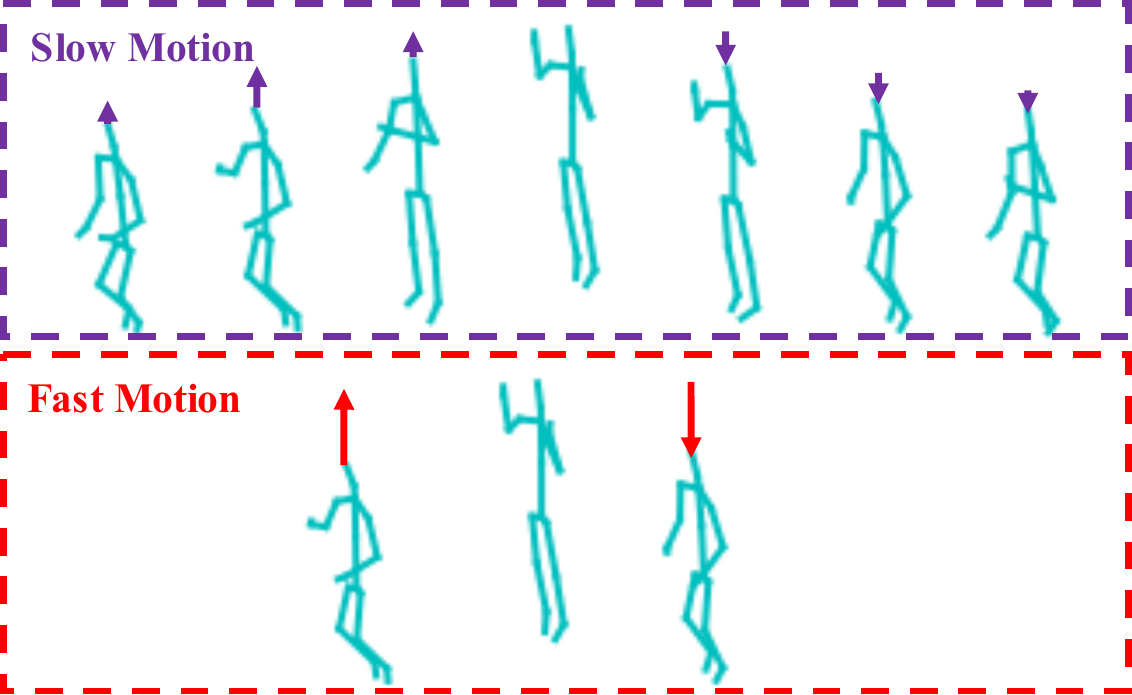}}
   \caption{Motion scale variation} 
\end{subfigure}
\begin{subfigure}[t]{.9\linewidth}
  {\includegraphics[width=.8\linewidth]{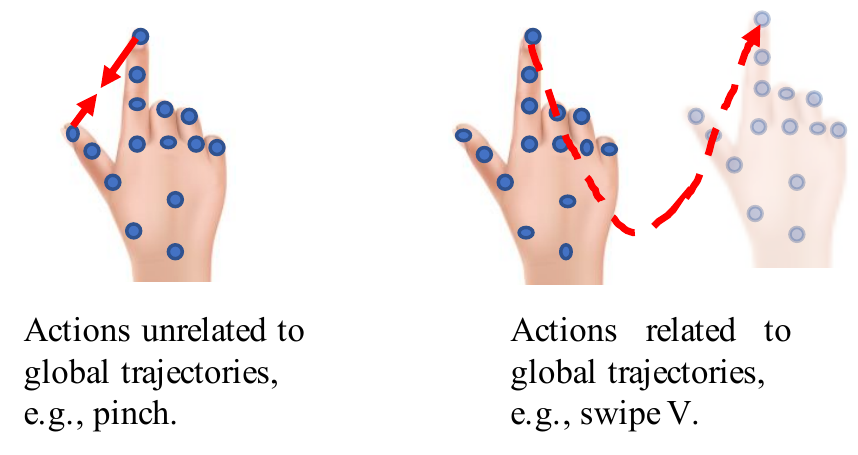}}
   \caption{Related/unrelated to global trajectories} 
\end{subfigure}
\begin{subfigure}[t]{.9\linewidth}
  {\includegraphics[width=.6\linewidth]{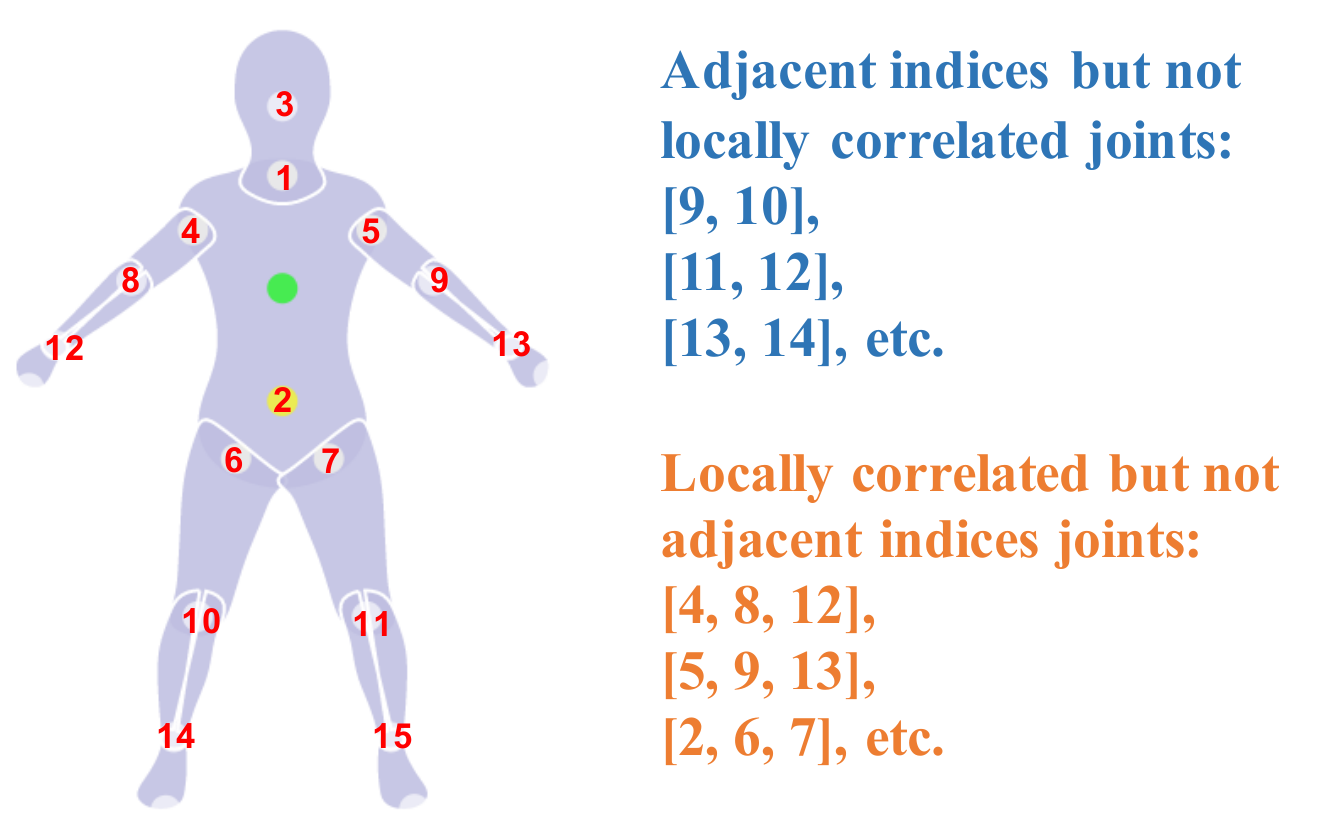}}
   \caption{Uncorrelated joint indices (PuppetModel~\cite{Jhuang:ICCV:2013})} 
\end{subfigure}
   \caption{Examples of skeleton sequence properties.}\label{fig:properties}
\end{figure}

More specifically, we conduct research on four types of skeleton sequence properties (see Fig.~\textcolor{red}{\ref{fig:properties}}): (a) location-viewpoint variation, (b) motion scale variation, (c) related/unrelated to global trajectories, (4) uncorrelated joint indices. To address challenges caused by these properties, previous works may prone to propose complicated neural network models, which end up with large model size. 

In contrast, we address these challenges by simplifying both the input feature and the network structure. Our JCD feature contains the location-viewpoint invariant information of skeleton sequences. Compared with other similar features, it can be easily computed and includes fewer elements. Since global motions cannot be incorporated into a location-viewpoint invariant feature, we introduce a two-scale global motion feature to improve the generalization of DD-Net. Besides, its two-scale structure makes it robust to the motion scale variance. Through an embedding process, DD-Net can automatically learn the proper correlation of joints, which is hard to be predefined by joint indices. 

Compared to methods relying on complicated model structures, DD-Net provides higher action recognition accuracy and demonstrates its generalization on our experiential datasets. With its efficiency both in terms of computational complexity and the number of parameters, DD-Net is sufficient to be applied in real applications.

\section{Related Works}

Nowadays, with the fast advancement of deep learning, skeleton acquisition is not limited to use motion capture systems~\cite{moeslund2001survey} and depth cameras~\cite{zhang2012microsoft}. The RGB data, for instance, can be used to infer 2D skeletons~\cite{cao2017realtime,xiao2018simple} or 3D skeletons~\cite{VNect_SIGGRAPH2017,ge2018real} in real time. Moreover, even WiFi signals can be used to estimate skeleton data~\cite{zhao2018through,wang2019can}. Those achievements have made skeleton-based action recognition available on a huge amount of multimedia resources and therefore have stimulated the model's development.

In general, in order to achieve a better performance for skeleton-based action recognition, previous studies attempt to work on two aspects: introduce new features for skeleton sequences~\cite{chen2011learning,de2016skeleton,liu2017enhanced,caputo20173,zhang2018fusing,choutas2018potion,chen2019mfa}, and, propose novel neural network architectures~\cite{Li2017Joint, lee2017ensemble, tang2018deep, yang2018action, devineau2018convolutional,hou2018spatial, ludl2019simple}. 

A good skeleton-sequence representation should contain global motion information and be location-viewpoint invariant. However, it is challenging to satisfy both requirements in one feature. The studies~\cite{de2016skeleton,caputo20173, choutas2018potion,chen2019mfa} focused on global motions without considering the location-viewpoint variation in their features. Other studies~\cite{chen2011learning,liu2017enhanced,zhang2018fusing}, on the contrary, introduced location-viewpoint invariant features without considering global motions. Our work bridges their gaps by seamlessly integrating a location-viewpoint invariant feature and a two-scale global motion feature together.    

Although Recurrent Neural Networks (RNNs) are commonly used in skeleton-based action recognition~\cite{du2015hierarchical,liu2016spatio,wang2017modeling, song2017end, li2017skeleton, zhang2018fusing}, we argue that it is relatively slow and difficult for parallel computing, compared with methods~\cite{Li2017Joint,devineau2018convolutional,chen2019mfa} that use Convolutional Neural Networks (CNNs). Since we take the model speed as one of our priorities, we utilize 1D CNNs to construct the backbone network of DD-Net.

\vspace{0.5cm}
\section{Methodology}

\begin{figure}[!htb]
\centering
  \includegraphics[width=1\columnwidth]{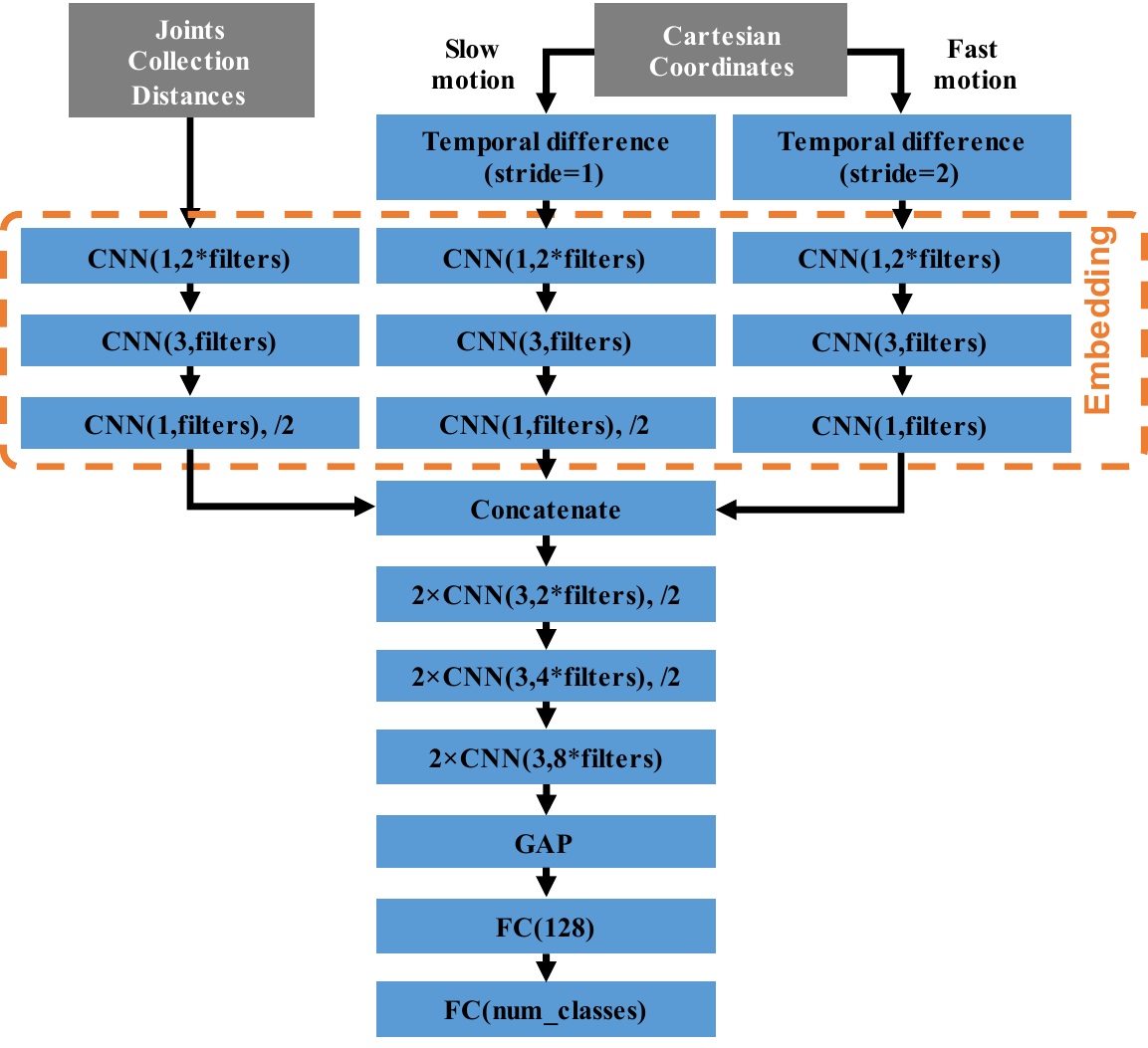}
  \caption{The network architecture of DD-Net. ``2$\times$CNN(3, 2*$filters$), /2'' denotes two 1D ConvNet layers (kernel size = 3, channels = 2*$filters$) and a Maxpooling (strides = 2). Other ConvNet layers are defined in the same format. GAP denotes Global Average Pooling. FC denotes Fully Connected Layers (or Dense Layers). We can change the model size by modifying $filters$.}
  \label{fig:DD-Net}
\end{figure}

The network architecture of Double-feature Double-motion Network (DD-Net) is shown in Fig.~\textcolor{red}{\ref{fig:DD-Net}}. In the following, we explain our motivation for designing input features and network structures of DD-Net.

\subsection{Modeling Location-viewpoint Invariant Feature by Joint Collection Distances (JCD)}

For skeleton-based action recognition, two types of input features are commonly used: the geometric feature~\cite{chen2011learning,zhang2018fusing} and the Cartesian coordinate feature~\cite{wang2017modeling, song2017end, yan2018spatial,hou2018spatial,zolfaghari2017chained}. The Cartesian coordinate feature is variant to locations and viewpoints. As Fig.~\textcolor{red}{\ref{fig:properties}} (a) shows, when skeletons are rotated or shifted, the Cartesian coordinate feature can be significantly changed. The geometric feature (\eg,~ angles/distances), on the other hand, is location-viewpoint invariant, and thereby it has been utilized for skeleton-based action recognition for a while. However, existing geometric features may need to be heavily redesigned from one dataset to another~\cite{chen2011learning,zhang2018fusing}, or, contain redundant elements~\cite{li2017skeleton}. To alleviate these issues, we introduce a Joint Collection Distances (JCD) feature.

\begin{figure}[!ht]
\centering
  \includegraphics[width=0.4\columnwidth]{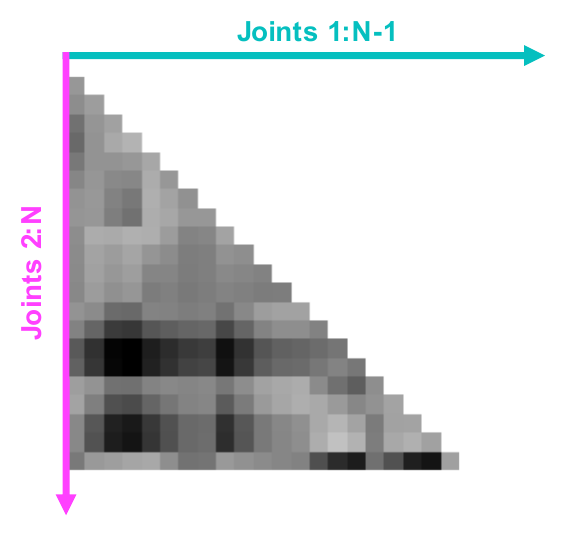}
  \caption{An example of Joint Collection Distances (JCD) feature at frame $k$, where the number of joints is $N$.}
  \label{fig:JCD}
\end{figure}

We calculate the Euclidean distances between a pair of collective joints to obtain a symmetric matrix. To reduce the redundancy, only the lower triangular matrix without the diagonal part is used as the JCD feature (see Fig.~\textcolor{red}{\ref{fig:JCD}}). Hence, the JCD feature is less than half the size of \cite{li2017skeleton}. 

In more detail, we assume the total frame number is $K$ ($K=32$ as the default setting) and there are totally $N$ joints for one subject. At frame $k$, the 3D Cartesian coordinates of joint $n$ is represented as $J_{i}^{k} = (x, y, z)$, while the 2D Cartesian coordinates is represented as $J_{i}^{k} = (x, y)$. Put all of joints together, we have a joint collection $S^{k} = \{J_{1}^{k}, J_{2}^{k}, . . . , J_{N}^{k} \}$. The formula for calculating the JCD feature of $S^{k}$ is:
\begin{equation}
\centering
\begin{split}
    JCD^{k} =  
\begin{bmatrix}
& \left \|\overrightarrow{J_{2}^{k}J_{1}^{k}}\right\| & & &\\
 & \vdots   & \ddots & & \\
  & \vdots   & \cdots & \ddots & \\
  & \left \|\overrightarrow{J_{N}^{k}J_{1}^{k}}\right\| & \cdots &  \cdots & \left \|\overrightarrow{J_{N}^{k}J_{N-1}^{k}}\right\|
 \end{bmatrix};\\
   \end{split}
\end{equation}
where $\Big \|\overrightarrow{J_{i}^{k}J_{j}^{k}} \Big\| (i\neq j) $ denotes the Euclidean distance between $J_{i}^{k}$ and $J_{j}^{k}$. 

In our processing, the JCD feature is flattened to be a one-dimensional vector as our model's input. The dimension of flattened JCD is $\binom{N}{2}$. 

\subsection{Modeling Global Scale-invariant Motions by a Two-scale Motion Feature}

Although the JCD feature is location-viewpoint invariant, the same as other geometric features, it does not contain global motion information. When actions are associated with global trajectories (see Fig.~\textcolor{red}{\ref{fig:properties}} (c)), solely using the JCD feature is insufficient. Unlike previous works that only utilize either the geometric feature~\cite{chen2011learning,zhang2018fusing} or the Cartesian coordinate feature~\cite{Li2017Joint, lee2017ensemble, tang2018deep, yang2018action}, our DD-Net seamlessly integrates both of them. 

We calculate the temporal differences (\ie,~the speed) of the Cartesian coordinate feature to obtain global motions, which is location-invariant. For the same action, however, the scale of global motions may not be exactly identical. Some might be faster, and others might be slower (see Fig.~\textcolor{red}{\ref{fig:properties}} (b)). To learn a robust global motion feature, both fast and slow motions should be considered. Conferring this intuition to DD-Net, we employ a fast global motion and a slow global motion to form a two-scale global motion feature. This idea is inspired by the two-scale optical flows proposed for RGB-based action recognition~\cite{feichtenhofer2018slowfast}.

Technically, the two-scale motions can be generated by the following equation:
\begin{equation}
\begin{split}
    M_{slow}^{k} = S^{k+1} - S^{k}, k \in \{1,2,3,...,K-1\};\\
    M_{fast}^{k} = S^{k+2} - S^{k}, k \in \{1,3,...,K-2\};
\end{split}
\end{equation}
where $M_{slow}^{k}$ and $M_{fast}^{k}$ denote the slow motion and the fast motion at frame $k$, respectively. $S^{k+1}$ and $S^{k+2}$ are behind the $S^{k}$ of one frame and two frames, respectively. Corresponding to $S^{[1,...,K]}$, we have $M_{slow}^{[1,...,K-1]}$ and $M_{fast}^{[1,...,K/2-1]}$.

To generate an one-dimensional input at each frame, we reshape $M_{slow}^{k}$ and $M_{fast}^{k}$ as $M_{slow}^{k} \in \mathbb{R}^{D_{motion}}$ and $M_{fast}^{k} \in \mathbb{R}^{D_{motion}}$, respectively, where $D_{motion}$ is the dimension of flattened vector. To match the frame number of the JCD feature, we perform linear interpolation to resize $M_{slow}^{[1,...,K-1]}$ and $M_{fast}^{[1,...,K/2-1]}$ as $M_{slow}^{[1,...,K]}$ and $M_{fast}^{[1,...,K/2]}$, respectively. Consequently, two-scale global motion feature is composed of $M_{slow}^{[1,...,K]} \in \mathbb{R}^{K \times D_{motion}}$ and $M_{fast}^{[1,...,K/2]} \in \mathbb{R}^{(K/2) \times D_{motion}}$. Such a process can be done in our DD-Net, and only the Cartesian coordinate feature is needed as the input.

\subsection{Modeling Joint Correlations by an Embedding}

Fig.~\textcolor{red}{\ref{fig:properties}} (d) shows that the joint indices (\ie,~the IDs of the head, left and right hands, \etc) are not locally correlated. Moreover, in different actions, the correlation of joints could be dynamically changed. Hence, the difficulty arises when we try to pre-define the correlation of joints by manually ordering their indices.  

Since most of neural networks inherently assume that inputs are locally correlated, directly processing the locally uncorrelated joint feature is inappropriate. To tackle this problem, our DD-Net embeds the JCD feature and the two-scale motion feature into latent vectors at each frame. The correlation of joints is automatically learned through the embedding. As another benefit, the embedding process also reduces the effect of skeleton noise. 

More formally, let embedding representations of $JCD^{k}$, $M_{slow}^{k}$ and $M_{fast}^{k}$ to be  $\varepsilon_{JCD}^{k}$, $\varepsilon_{M_{slow}}^{k}$ and $\varepsilon_{M_{fast}}^{k}$, respectively, the embedding operation is as follows,
\begin{equation}
\begin{split}
    \varepsilon_{JCD}^{k} = \mathit{Embed_{1}}(JCD^{k});\\
    \varepsilon_{M_{slow}}^{k} = \mathit{Embed_{1}}(M_{slow}^{k});\\
    \varepsilon_{M_{fast}}^{k} = \mathit{Embed_{2}}(M_{fast}^{k}).
\end{split}
\end{equation}
where the $\mathit{Embed_{1}}$ is defined as $Conv1D(1,2*filters) \rightarrow Conv1D(3,\newline filters) \rightarrow Conv1D(1,filters)$, and the $\mathit{Embed_{2}}$ is defined as $\newline Conv1D(1,2*filters) \rightarrow Conv1D(3,filters) \rightarrow Conv1D(1,filters) \newline \rightarrow Maxpooling(2)$, because $JCD^{k}$ and $M_{slow}^{k}$ have double the temporal length of $M_{fast}^{k}$.

DD-Net futher concatenates embedding features to a representation $\varepsilon^{k}$ by
\begin{equation}
\begin{split}
    \varepsilon^{k} = \varepsilon_{JCD}^{k} \oplus  \varepsilon_{M_{slow}}^{k} \oplus \varepsilon_{M_{fast}}^{k},\\
    \wrt~~\varepsilon^{k} \in \mathbb{R}^{(K/2) \times filters};
\end{split}
\end{equation}
where $\oplus$ is the concatenation operation.

After the embedding process, subsequent processes are not affected by the joint indices, and therefore DD-Net can use the 1D ConvNet to learn the temporal information as Fig.~\textcolor{red}{\ref{fig:DD-Net}} shows.

\vspace{0.5cm}
\section{Experiments}

\subsection{Experimental Datasets}

We select two skeleton-based action recognition datasets, as SHREC dataset~\cite{de2017shrec} and JHMDB dataset~\cite{Jhuang:ICCV:2013}, to evaluate our DD-Net from different perspectives (see Table \textcolor{red}{\ref{Table:datasets}}). 

\begin{table}[!h]
\renewcommand{\arraystretch}{1.3}
\caption{Properties of experimental datasets}
\label{Table:datasets}
\centering
\scalebox{0.92}{
\begin{tabular}{c|c|c}
\hline
& \textbf{SHREC} & \textbf{JHMDB} \\
& \textbf{Dataset} & \textbf{Dataset} \\\hline
\textbf{Number of samples} &2,800 &928 \\\hline
\textbf{Training/} &1 Training Set & 3 Split Training/ \\
\textbf{Testing Setup} &1 Testing Set & Testing Sets\\\hline
 \textbf{Dimension of skeletons} &3  &2 \\\hline
\textbf{subject}  &hand  & body \\\hline
\textbf{Number of actions} &14 and 28  &21 \\\hline
\textbf{Actions are strongly} &&\\
\textbf{correlated to}   & \cmark & \xmark\\
\textbf{global trajectories} &&\\\hline
\end{tabular}}
\end{table}

Although other information (\eg,~ RGB data) is available, only the skeleton information is used in our experiments. 3D skeletons are given by SHREC dataset, which are derived from RGB-D data and contain more spatial information. In JHMDB dataset, 2D skeletons are interpreted from RGB videos, which can be applied in more general cases where inferring the depth information may be hard or impossible. Besides, actions in SHREC dataset are strongly correlated to the subject's global trajectories (\eg,~a hand swipes a `V' shape), while JHMDB dataset may have a weak connection with global trajectories. We show how these properties affect the performance and demonstrate the generalization of DD-Net in our ablation studies.

\subsection{Evaluation Setup}

The SHREC dataset is evaluated in two cases: 14 gestures and 28 gestures. The JHMDB dataset is evaluated by using the manually annotated skeletons, and we average the results from three split training/testing sets.

In ablation studies, we explore how each DD-Net component contributes to the action recognition performance by removing one component while remaining others unchanged. Furthermore, we also explore how the performance varies with different model size by adjusting the value of $filters$ in Fig.~\textcolor{red}{\ref{fig:DD-Net}}.

\subsection{Implementation Details}
Since the DD-Net is small, it is feasible to put all of the training sets into one batch on a single GTX 1080Ti GPU. We choose Adam ($\beta_1=0.9$, $\beta_2=0.999$) \cite{kingma2014adam} as the optimizer, with an annealing learning rate that drops from $1^{-3}$ to $1^{-5}$. During the training, DD-Net only takes a temporal augmentation, which randomly selects $0.9$ of entire frames.

To demonstrate the superiority of DD-Net, we do not apply any ensemble strategy or pre-trained weights to boost the performance. To make DD-Net can be easily deployed to real applications, we implement it by Keras \cite{chollet2015keras} backend in Tensorflow, which is ``notorious'' for its slow execution speed. Using other neural network frameworks may make DD-Net faster.

\subsection{Result Analysis and Discussion}

\begin{figure}[!htb]
\centering
  \includegraphics[width=0.9\columnwidth]{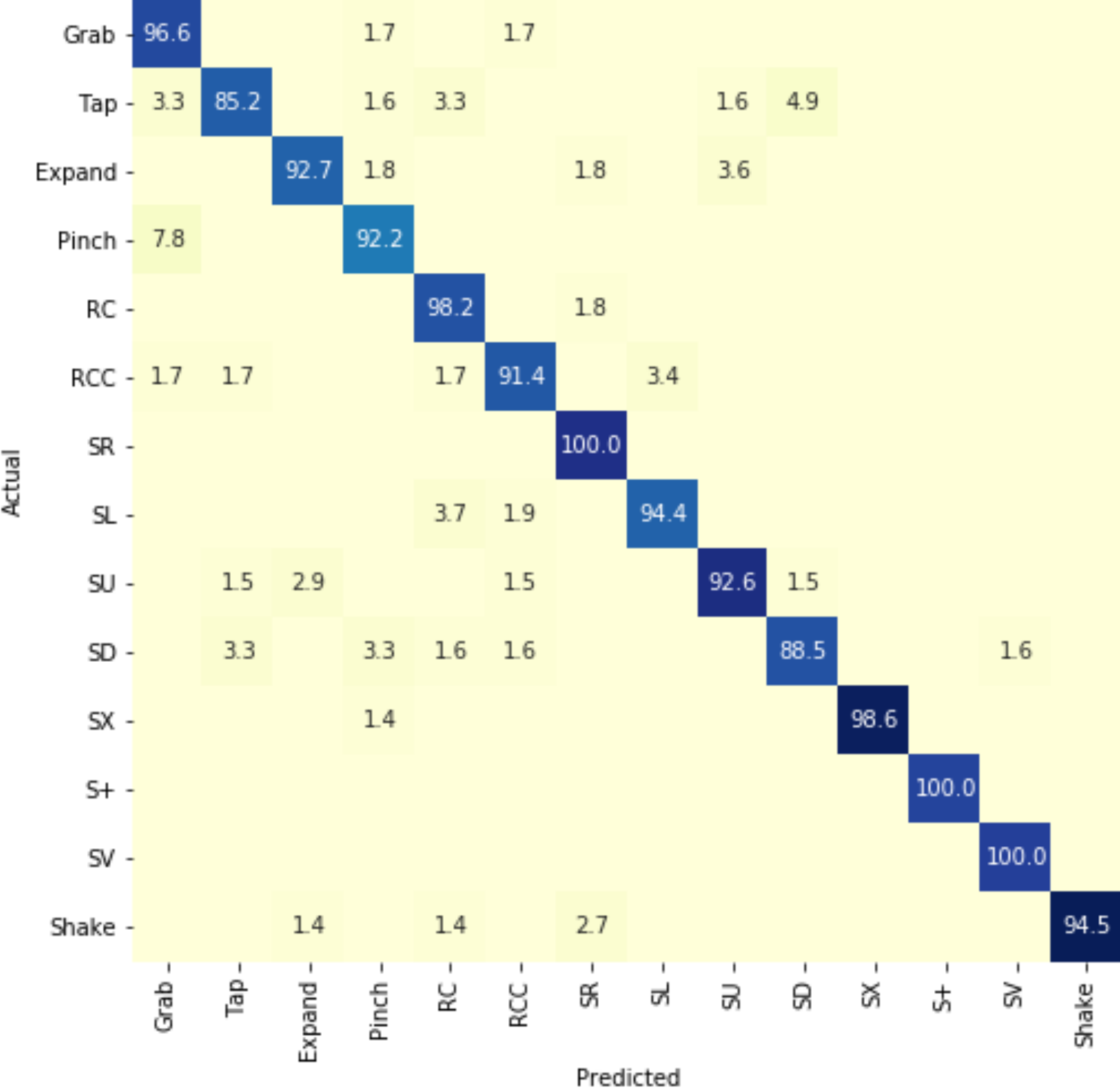}
  \caption{Confusion matrix of SHREC dataset (14 hand actions) obtained by DD-Net.}
  \label{fig:cm_14}
\end{figure}

\begin{figure*}[!htb]
\centering
  \includegraphics[width=0.66\linewidth]{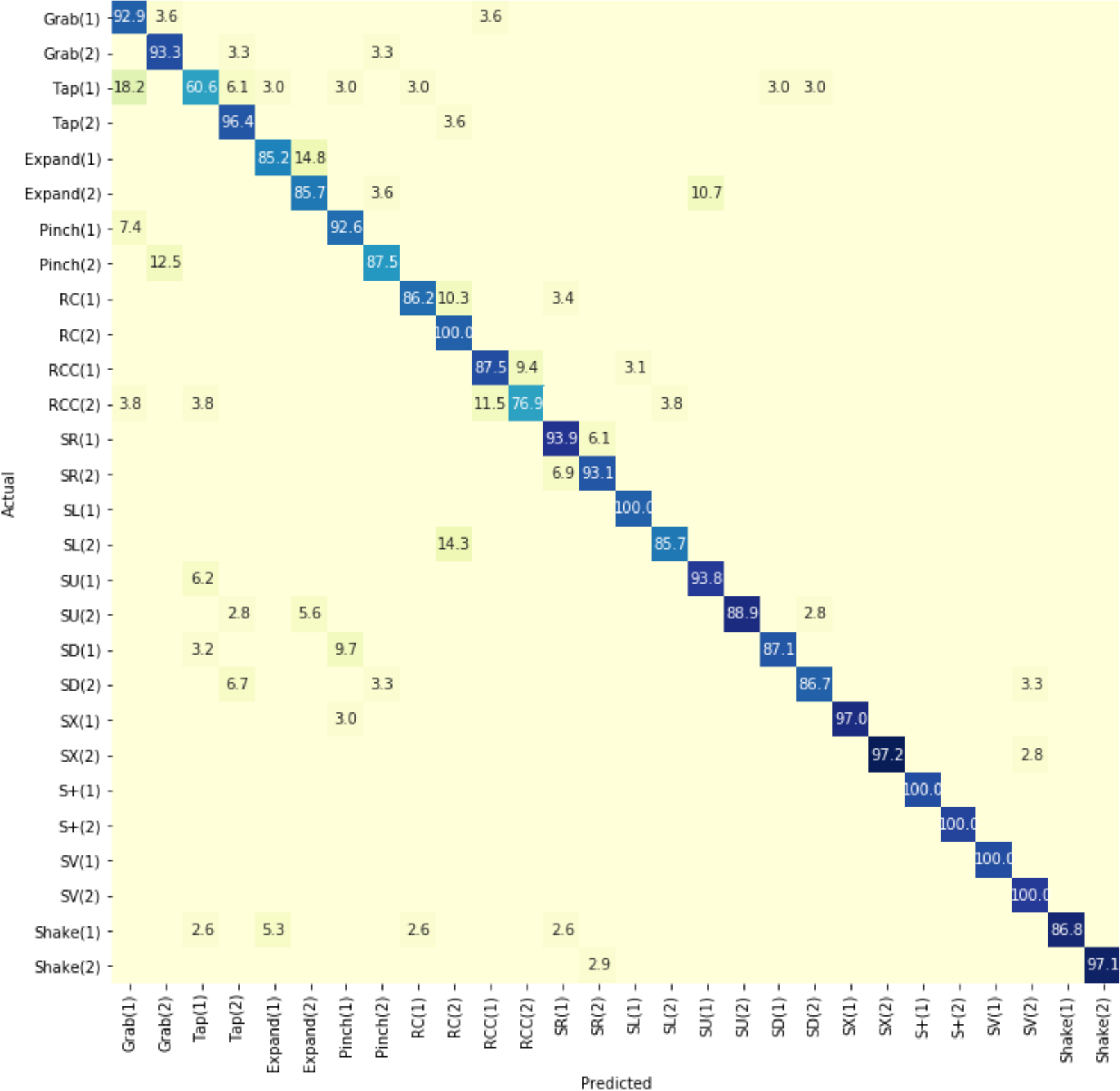}
  \caption{Confusion matrix of SHREC dataset (28 hand actions) obtained by DD-Net.}
  \label{fig:cm_28}
\end{figure*}

The action recognition results of SHREC dataset are presented in Table~\textcolor{red}{\ref{table:SHREC}} and more details are listed in their confusion matrix. The confusion matrix of $14$ actions and $28$ actions are  Fig.~\textcolor{red}{\ref{fig:cm_14}} and Fig.~\textcolor{red}{\ref{fig:cm_28}}, respectively. The action recognition results of JHMDB dataset are presented in Table~\textcolor{red}{\ref{table:JHMDB}}.

\tabcolsep=5pt
\begin{table}[!h]
\renewcommand{\arraystretch}{1.3}
\caption{Results on SHREC (Using 3D skeletons only) \cite{de2017shrec}}
\label{table:SHREC}
\centering
\scalebox{0.80}{
\begin{tabular}{c|c|c|c|c}
\hline
\textbf{Methods} & \textbf{Parameters} & \textbf{14}   & \textbf{28} & \textbf{Speed} \\
& & \textbf{Gestures}   & \textbf{ Gestures}& \textbf{ on GPU}\\\hline
Dynamic hand \cite{de2016skeleton} (CVPRW16)&- &88.2\% &81.9\% &-\\
Key-frame CNN \cite{de2017shrec} (3DOR17)&7.92 M&82.9\%&71.9\% &-\\
3 Cent \cite{caputo20173} (STAG17) &-&77.9\%&- &-\\
CNN+LSTM\cite{nunez2018convolutional} (PR18) &8-9 M &89.8\%& 86.3\% &238 FPS\\
Parallel CNN \cite{devineau2018convolutional} (RFIAP18) &13.83 M &91.3\%& 84.4\% &-\\
STA-Res-TCN \cite{hou2018spatial} (Gesture18) & 5-6 M&93.6\%& 90.7\% & 303 FPS\\
MFA-Net \cite{chen2019mfa} (Sensor19) & - & 91.3\%& 86.6\% &361 FPS\\\hline
DD-Net (filters=64, w/o  &&&\\
global fast\&slow motion) & 1.70 M &55.2\%& 41.6\%&-\\
DD-Net (filters=64,&&&\\
w/o global slow motion)  & 1.76 M &92.7\%& 90.2\%&-\\
DD-Net (filters=64,&&&\\
w/o global fast motion)  & 1.76 M &93.3\%& 90.5\%&-\\
DD-Net (filters=64) &1.82 M &\textbf{94.6\%}&\textbf{ 91.9\%} &2,200 FPS\\
DD-Net (filters=32)  & 0.50 M &93.5\%& 90.4\% &3,100 FPS\\
DD-Net (filters=16)  & \textbf{0.15 M} &91.8\%& 90.0\% &3,500 FPS\\\hline
\end{tabular}}
\end{table}

\tabcolsep=5pt
\begin{table}[!h]
\renewcommand{\arraystretch}{1.3}
\caption{Results on JHMDB (Using 2D skeletons only) \cite{Jhuang:ICCV:2013}}
\label{table:JHMDB}
\centering
\scalebox{0.9}{
\begin{tabular}{c|c|c|c}
\hline
\textbf{Methods} & \textbf{Parameters} & \textbf{Manually} & \textbf{Speed}\\
&&\textbf{annotated} &\textbf{on GPU}\\
&&\textbf{skeletons} &\\ \hline
 Chained Net \cite{zolfaghari2017chained} (ICCV17)&17.50 M &56.8\% & 33 FPS \\
 EHPI \cite{ludl2019simple} (ITSC19) &1.22 M &65.5\% & 29 FPS\\
  PoTion \cite{choutas2018potion} (CVPR18)&4.87 M &67.9\% & 100 FPS \\\hline
DD-Net (filters=64, w/o  & &\\
global fast\&slow motion) & 0.46 M &70.3\%&-\\
DD-Net (filters=64,& &\\
w/o global slow motion)  & 0.48 M &72.5\% &-\\
DD-Net (filters=64,& &\\
w/o global fast motion)  & 0.48 M &73.1\% &-\\
DD-Net (filters=64) &1.82 M &\textbf{77.2}\% &2,200 FPS\\
DD-Net (filters=32) &0.50 M &73.7\%&3,100 FPS\\
DD-Net (filters=16) &\textbf{0.15 M} &65.7\% &3,500 FPS\\\hline
\end{tabular}}
\end{table}

Overall, although DD-Net takes fewer parameters, it can achieve superior results on SHREC dataset and JHMDB dataset. The confusion matrix also shows that DD-Net is robust to each action class. Despite the data property divergence existing, DD-Net demonstrates its generalization ability, which suggests it can accommodate a wide range of skeleton-based action recognition scenarios.

From ablation studies, we can inspect that when actions are strongly correlated to global trajectories (\eg,~ SHREC dataset), just using the JCD feature cannot produce a satisfactory performance. When actions are not strongly correlated to global trajectories (\eg,~ JHMDB dataset), the global motion feature still helps to improve the performance, but not as significant as the previous case. Such results agree with our assumptions: although the JCD feature is location-viewpoint invariant, it is isolated from global motions. The results also show that using the two-scale motion feature generates higher classification accuracy than only using a one-scale motion feature, which suggests that our proposed two-scale global motion feature is more robust to scale variation of motions. With the same components, DD-Net can adjust its model size by modifying the value of $filters$ in CNN layers. We select $64$, $32$ and $16$ as the values of $filters$ to perform experiments. When DD-Net reaches the best performance on SHREC and JHMDB datasets, the values of $filters$ are $64$. It is worth noting that DD-Net can generate comparable results by only using \emph{$0.15$ million} parameters.

In addition, since DD-net employs one-dimensional CNNs to extract the feature, it is much faster than other models that use RNNs \cite{wang2017modeling,zhang2018fusing,song2017end,lee2017ensemble} or 2D/3D CNNs \cite{ devineau2018convolutional,liu2018recognizing,zolfaghari2017chained,choutas2018potion, ludl2019simple}. During its inferences, DD-Net's speed can reach around 3,500 FPS on one GPU (\ie,~GTX 1080Ti), or, 2,000 FPS on one CPU (\ie,~Intel E5-2620). While RNN-based models face great challenges for parallel processing (due to sequential dependency), our DD-Net does not have this issue because CNNs are used. Therefore, whether low-computational (\eg,~ on small devices) or high-computational applications (\eg,~ on parallel computing stations) are concerned, our DD-Net enjoys significant superiority.

\section{Conclusion}
By analyzing the basic properties of skeleton sequence, we propose two features and a DD-Net for efficient skeleton-based action recognition. Although DD-Net only contains a few parameters, it can achieve state-of-the-art performance on our experimental datasets. Due to the simplicity of DD-Net, many possibilities exist to enhance/extend it for broader studies. For instance, online action recognition can be approached by modifying the frame sampling strategies; RGB data or depth data could be used with it to further improve the action recognition performance; it is also possible to extend it for temporal action detection by adding temporal segmentation related modules.

\section{ACKNOWLEDGEMENTS}
Part of this work was supported by JSPS KAKENHI Grant Numbers JP17H06101 and JP17K00237, the Royal Society under IEC \textbackslash R3 \textbackslash \ 170013 - International Exchanges 2017 Cost Share (Japan and Taiwan only), and a MSRA Collaborative Research 2019 Grant by Microsoft Research Asia.

\ifCLASSOPTIONcaptionsoff
  \newpage
\fi

\bibliographystyle{IEEEtran}
\bibliography{ref}

\begin{thebibliography}{10}
\providecommand{\url}[1]{#1}
\csname url@samestyle\endcsname
\providecommand{\newblock}{\relax}
\providecommand{\bibinfo}[2]{#2}
\providecommand{\BIBentrySTDinterwordspacing}{\spaceskip=0pt\relax}
\providecommand{\BIBentryALTinterwordstretchfactor}{4}
\providecommand{\BIBentryALTinterwordspacing}{\spaceskip=\fontdimen2\font plus
\BIBentryALTinterwordstretchfactor\fontdimen3\font minus
  \fontdimen4\font\relax}
\providecommand{\BIBforeignlanguage}[2]{{%
\expandafter\ifx\csname l@#1\endcsname\relax
\typeout{** WARNING: IEEEtran.bst: No hyphenation pattern has been}%
\typeout{** loaded for the language `#1'. Using the pattern for}%
\typeout{** the default language instead.}%
\else
\language=\csname l@#1\endcsname
\fi
#2}}
\providecommand{\BIBdecl}{\relax}
\BIBdecl

\bibitem{ren2011robust}
Z.~Ren, J.~Meng, J.~Yuan, and Z.~Zhang, ``Robust hand gesture recognition with
  kinect sensor,'' in \emph{Proceedings of the 19th ACM international
  conference on Multimedia}.\hskip 1em plus 0.5em minus 0.4em\relax ACM, 2011,
  pp. 759--760.

\bibitem{wei2014skeleton}
S.-E. Wei, N.~C. Tang, Y.-Y. Lin, M.-F. Weng, and H.-Y.~M. Liao,
  ``Skeleton-augmented human action understanding by learning with
  progressively refined data,'' in \emph{Proceedings of the 1st ACM
  International Workshop on Human Centered Event Understanding from
  Multimedia}.\hskip 1em plus 0.5em minus 0.4em\relax ACM, 2014, pp. 7--10.

\bibitem{chang2011kinect}
Y.-J. Chang, S.-F. Chen, and J.-D. Huang, ``A kinect-based system for physical
  rehabilitation: A pilot study for young adults with motor disabilities,''
  \emph{Research in developmental disabilities}, vol.~32, no.~6, pp.
  2566--2570, 2011.

\bibitem{de2017shrec}
Q.~De~Smedt, H.~Wannous, J.-P. Vandeborre, J.~Guerry, B.~Le~Saux, and
  D.~Filliat, ``Shrec'17 track: 3d hand gesture recognition using a depth and
  skeletal dataset,'' in \emph{10th Eurographics Workshop on 3D Object
  Retrieval}, 2017.

\bibitem{devineau2018convolutional}
G.~Devineau, W.~Xi, F.~Moutarde, and J.~Yang, ``Convolutional neural networks
  for multivariate time series classification using both inter-and
  intra-channel parallel convolutions,'' in \emph{Reconnaissance des Formes,
  Image, Apprentissage et Perception (RFIAP'2018)}, 2018.

\bibitem{hou2018spatial}
J.~Hou, G.~Wang, X.~Chen, J.-H. Xue, R.~Zhu, and H.~Yang, ``Spatial-temporal
  attention res-tcn for skeleton-based dynamic hand gesture recognition,''
  \emph{gesture}, vol.~30, no.~5, p.~3, 2018.

\bibitem{zolfaghari2017chained}
M.~Zolfaghari, G.~L. Oliveira, N.~Sedaghat, and T.~Brox, ``Chained multi-stream
  networks exploiting pose, motion, and appearance for action classification
  and detection,'' in \emph{Computer Vision (ICCV), 2017 IEEE International
  Conference on Computer Vision}.\hskip 1em plus 0.5em minus 0.4em\relax IEEE,
  2017, pp. 2923--2932.

\bibitem{choutas2018potion}
V.~Choutas, P.~Weinzaepfel, J.~Revaud, and C.~Schmid, ``Potion: Pose motion
  representation for action recognition,'' in \emph{CVPR 2018}, 2018.

\bibitem{Jhuang:ICCV:2013}
H.~Jhuang, J.~Gall, S.~Zuffi, C.~Schmid, and M.~J. Black, ``Towards
  understanding action recognition,'' in \emph{International Conf. on Computer
  Vision (ICCV)}, Dec. 2013, pp. 3192--3199.

\bibitem{moeslund2001survey}
T.~B. Moeslund and E.~Granum, ``A survey of computer vision-based human motion
  capture,'' \emph{Computer vision and image understanding}, vol.~81, no.~3,
  pp. 231--268, 2001.

\bibitem{zhang2012microsoft}
Z.~Zhang, ``Microsoft kinect sensor and its effect,'' \emph{IEEE multimedia},
  vol.~19, no.~2, pp. 4--10, 2012.

\bibitem{cao2017realtime}
Z.~Cao, T.~Simon, S.-E. Wei, and Y.~Sheikh, ``Realtime multi-person 2d pose
  estimation using part affinity fields,'' in \emph{Computer Vision and Pattern
  Recognition (CVPR), 2017 IEEE Conference on}.\hskip 1em plus 0.5em minus
  0.4em\relax IEEE, 2017, pp. 1302--1310.

\bibitem{xiao2018simple}
B.~Xiao, H.~Wu, and Y.~Wei, ``Simple baselines for human pose estimation and
  tracking,'' in \emph{Proceedings of the European Conference on Computer
  Vision (ECCV)}, 2018, pp. 466--481.

\bibitem{VNect_SIGGRAPH2017}
\BIBentryALTinterwordspacing
D.~Mehta, S.~Sridhar, O.~Sotnychenko, H.~Rhodin, M.~Shafiei, H.-P. Seidel,
  W.~Xu, D.~Casas, and C.~Theobalt, ``Vnect: Real-time 3d human pose estimation
  with a single rgb camera,'' vol.~36, no.~4, 2017. [Online]. Available:
  \url{http://gvv.mpi-inf.mpg.de/projects/VNect/}
\BIBentrySTDinterwordspacing

\bibitem{ge2018real}
L.~Ge, H.~Liang, J.~Yuan, and D.~Thalmann, ``Real-time 3d hand pose estimation
  with 3d convolutional neural networks,'' \emph{IEEE Transactions on Pattern
  Analysis and Machine Intelligence}, 2018.

\bibitem{zhao2018through}
M.~Zhao, T.~Li, M.~Abu~Alsheikh, Y.~Tian, H.~Zhao, A.~Torralba, and D.~Katabi,
  ``Through-wall human pose estimation using radio signals,'' in
  \emph{Proceedings of the IEEE Conference on Computer Vision and Pattern
  Recognition}, 2018, pp. 7356--7365.

\bibitem{wang2019can}
F.~Wang, S.~Panev, Z.~Dai, J.~Han, and D.~Huang, ``Can wifi estimate person
  pose?'' \emph{arXiv preprint arXiv:1904.00277}, 2019.

\bibitem{chen2011learning}
C.~Chen, Y.~Zhuang, F.~Nie, Y.~Yang, F.~Wu, and J.~Xiao, ``Learning a 3d human
  pose distance metric from geometric pose descriptor,'' \emph{IEEE
  Transactions on Visualization and Computer Graphics}, vol.~17, no.~11, pp.
  1676--1689, 2011.

\bibitem{de2016skeleton}
Q.~De~Smedt, H.~Wannous, and J.-P. Vandeborre, ``Skeleton-based dynamic hand
  gesture recognition,'' in \emph{Proceedings of the IEEE Conference on
  Computer Vision and Pattern Recognition Workshops}, 2016, pp. 1--9.

\bibitem{liu2017enhanced}
M.~Liu, H.~Liu, and C.~Chen, ``Enhanced skeleton visualization for view
  invariant human action recognition,'' \emph{Pattern Recognition}, vol.~68,
  pp. 346--362, 2017.

\bibitem{caputo20173}
F.~M. Caputo, P.~Prebianca, A.~Carcangiu, L.~D. Spano, and A.~Giachetti, ``A 3
  cent recognizer: Simple and effective retrieval and classification of mid-air
  gestures from single 3d traces,'' \emph{Smart Tools and Apps for Graphics.
  Eurographics Association}, 2017.

\bibitem{zhang2018fusing}
S.~Zhang, Y.~Yang, J.~Xiao, X.~Liu, Y.~Yang, D.~Xie, and Y.~Zhuang, ``Fusing
  geometric features for skeleton-based action recognition using multilayer
  lstm networks,'' \emph{IEEE Transactions on Multimedia}, vol.~20, no.~9, pp.
  2330--2343, 2018.

\bibitem{chen2019mfa}
X.~Chen, G.~Wang, H.~Guo, C.~Zhang, H.~Wang, and L.~Zhang, ``Mfa-net: Motion
  feature augmented network for dynamic hand gesture recognition from skeletal
  data,'' \emph{Sensors}, vol.~19, no.~2, p. 239, 2019.

\bibitem{Li2017Joint}
C.~Li, Y.~Hou, P.~Wang, and W.~Li, ``Joint distance maps based action
  recognition with convolutional neural network,'' \emph{IEEE Signal Processing
  Letters}, vol.~24, no.~5, pp. 624--628, 2017.

\bibitem{lee2017ensemble}
I.~Lee, D.~Kim, S.~Kang, and S.~Lee, ``Ensemble deep learning for
  skeleton-based action recognition using temporal sliding lstm networks,'' in
  \emph{Proceedings of the IEEE International Conference on Computer Vision},
  2017, pp. 1012--1020.

\bibitem{tang2018deep}
Y.~Tang, Y.~Tian, J.~Lu, P.~Li, and J.~Zhou, ``Deep progressive reinforcement
  learning for skeleton-based action recognition,'' in \emph{Proceedings of the
  IEEE Conference on Computer Vision and Pattern Recognition}, 2018, pp.
  5323--5332.

\bibitem{yang2018action}
Z.~Yang, Y.~Li, J.~Yang, and J.~Luo, ``Action recognition with spatio-temporal
  visual attention on skeleton image sequences,'' \emph{IEEE Transactions on
  Circuits and Systems for Video Technology}, 2018.

\bibitem{ludl2019simple}
D.~Ludl, T.~Gulde, and C.~Curio, ``Simple yet efficient real-time pose-based
  action recognition,'' \emph{arXiv preprint arXiv:1904.09140}, 2019.

\bibitem{du2015hierarchical}
Y.~Du, W.~Wang, and L.~Wang, ``Hierarchical recurrent neural network for
  skeleton based action recognition,'' in \emph{Proceedings of the IEEE
  conference on computer vision and pattern recognition}, 2015, pp. 1110--1118.

\bibitem{liu2016spatio}
J.~Liu, A.~Shahroudy, D.~Xu, and G.~Wang, ``Spatio-temporal lstm with trust
  gates for 3d human action recognition,'' in \emph{European Conference on
  Computer Vision}.\hskip 1em plus 0.5em minus 0.4em\relax Springer, 2016, pp.
  816--833.

\bibitem{wang2017modeling}
H.~Wang and L.~Wang, ``Modeling temporal dynamics and spatial configurations of
  actions using two-stream recurrent neural networks,'' in \emph{e Conference
  on Computer Vision and Pa ern Recognition (CVPR)}, 2017.

\bibitem{song2017end}
S.~Song, C.~Lan, J.~Xing, W.~Zeng, and J.~Liu, ``An end-to-end spatio-temporal
  attention model for human action recognition from skeleton data.'' in
  \emph{AAAI}, vol.~1, no.~2, 2017, pp. 4263--4270.

\bibitem{li2017skeleton}
C.~Li, P.~Wang, S.~Wang, Y.~Hou, and W.~Li, ``Skeleton-based action recognition
  using lstm and cnn,'' in \emph{Multimedia \& Expo Workshops (ICMEW), 2017
  IEEE International Conference on}.\hskip 1em plus 0.5em minus 0.4em\relax
  IEEE, 2017, pp. 585--590.

\bibitem{yan2018spatial}
S.~Yan, Y.~Xiong, and D.~Lin, ``Spatial temporal graph convolutional networks
  for skeleton-based action recognition,'' \emph{arXiv preprint
  arXiv:1801.07455}, 2018.

\bibitem{feichtenhofer2018slowfast}
C.~Feichtenhofer, H.~Fan, J.~Malik, and K.~He, ``Slowfast networks for video
  recognition,'' \emph{arXiv preprint arXiv:1812.03982}, 2018.

\bibitem{kingma2014adam}
D.~P. Kingma and J.~Ba, ``Adam: A method for stochastic optimization,''
  \emph{arXiv preprint arXiv:1412.6980}, 2014.

\bibitem{chollet2015keras}
F.~Chollet \emph{et~al.}, ``Keras,'' 2015.

\bibitem{nunez2018convolutional}
J.~C. Nunez, R.~Cabido, J.~J. Pantrigo, A.~S. Montemayor, and J.~F. Velez,
  ``Convolutional neural networks and long short-term memory for skeleton-based
  human activity and hand gesture recognition,'' \emph{Pattern Recognition},
  vol.~76, pp. 80--94, 2018.

\bibitem{liu2018recognizing}
M.~Liu and J.~Yuan, ``Recognizing human actions as the evolution of pose
  estimation maps,'' in \emph{Proceedings of the IEEE Conference on Computer
  Vision and Pattern Recognition}, 2018, pp. 1159--1168.

\end{thebibliography}

\end{document}